\documentclass[runningheads]{llncs}
\usepackage[T1]{fontenc}
\usepackage{graphicx}

\usepackage[utf8]{inputenc}
\usepackage[small]{caption}
\usepackage{subcaption}
\usepackage{graphicx}
\usepackage{tabularx}
\usepackage[caption=false]{subfig}
\usepackage{amsmath}
\usepackage{mathtools}
\usepackage{array}
\usepackage{booktabs}
\usepackage{subcaption}
\usepackage{multirow}
\usepackage{siunitx}
\begin{document}
\title{Improving performance of heart rate time series classification by grouping subjects\thanks{First and second author contributed equally.}}
% \title{Contribution Title\thanks{Supported by organization x.}}
%
\titlerunning{Improving classification performance by grouping}
\author{Michael Beekhuizen\inst{1}\orcidID{0009-0000-3874-6527}
Arman Naseri\inst{1, 2}\orcidID{0000-0002-5676-6596}
David Tax\inst{1}\orcidID{0000-0002-5153-9087}
Ivo van der Bilt\inst{2}\orcidID{0000-0002-1151-2461}
Marcel Reinders\inst{1}\orcidID{0000-0002-1148-1562}}
\authorrunning{M. Beekhuizen, A. Naseri, et al.}
\institute{Pattern Recognition and Bioinformatics, Delft University of Technology, Delft, The Netherlands
\\ \and
Department of Cardiology, Haga Teaching Hospital, The Hague, The Netherlands\\
}
\maketitle              % typeset the header of the contribution
\begin{abstract}
Unlike the more commonly analyzed ECG or PPG data for activity classification, heart rate time series data is less detailed, often noisier and can contain missing data points. Using the BigIdeasLab\_STEP dataset, which includes heart rate time series annotated with specific tasks performed by individuals, we sought to determine if general classification was achievable.

Our analyses showed that the accuracy is sensitive to the choice of window/stride size. Moreover, we found variable classification performances between subjects due to differences in the physical structure of their hearts. Various techniques were used to minimize this variability. First of all, normalization proved to be a crucial step and significantly improved the performance. Secondly, grouping subjects and performing classification inside a group helped to improve performance and decrease inter-subject variability. Finally, we show that including handcrafted features as input to a deep learning (DL) network improves the classification performance further.  

Together, these findings indicate that heart rate time series can be utilized for classification tasks like predicting activity. However, normalization or grouping techniques need to be chosen carefully to minimize the issue of subject variability.

\end{abstract}
\section{Introduction}
In recent years, wearable devices and smartwatches have been equipped with more sensors, including electrocardiogram (ECG) and photoplethysmography (PPG) sensors, for the estimation of heart rate and heart rhythm \cite{RN31}. These developments enable us to collect long-term heart rate time series data of a subject's heart rate in beats per minute (BPM). In the research community, there are many papers that attempt to perform classification using ECG or PPG data.
While ECG and PPG data shows each heartbeat's characteristics in detail, heart rate data summarizes this based on the time elapsed between heartbeats. For heart rate, we receive a single measurement, representing beats per minute, at regular intervals—like once every few seconds. This is therefore a more challenging signal to perform classification tasks on.
Research on the analysis and usage of heart rate time series has been performed for example for cardiovascular risk detection\cite{RN4}\cite{RN130}) and sleep analysis \cite{RN113}.
In this paper, we will look into the classification of heart rate time series data to predict different activities a subject is doing. This is interesting to test because it would imply that we can use heart rate data in the future for more complex classification problems, like heart disease detection. We will make use of the BigIdeasLab\_STEP\cite{BigIdeasLabDS} dataset which contains annotated heart rate time series of subjects performing different activities. 

Magure et al. perform activity classification\cite{RN131} which was mostly possible due to the fact that the accelerometer was placed at strategic places to identify specific movements and the subjects all had the same age and fitness level. In contrast, Bent and Dunn\cite{BigIdeasLabDS} conducted a study involving subjects of varying skin tones performing different physical activities while wearing multiple smartwatches. Their findings revealed no statistically significant difference in accuracy across skin tones. However, there were notable increases in error during physical activity compared to rest. Specifically, the absolute error during physical activity was on average 30\% greater than during periods of rest.

However, constructing a unified classification model is challenging due to the diverse characteristics between devices and subjects. Therefore we propose to group similar subjects together and construct a model for each group.

% \section{Related works}
% \label{sec:relatedworks}

%  The next section describes the experiment and the results we achieved on the classification of the heart rate times series. 

\renewcommand\tabularxcolumn[1]{m{#1}}
\newcolumntype{Z}{>{\raggedleft\arraybackslash}X}
\newcolumntype{Y}{>{\centering\arraybackslash}X}

\section{Results}
\label{sec:results}

The BigIdeasLab\_STEP dataset contains around 13 minutes of heart rate time series data per subject. This dataset is annotated with the activity a subject is performing. The activities were: resting, breathing, performing an activity (walking), resting after the activity and typing. The data is split up into windows of fixed size and a specific stride is used between each window. A window size refers to the number of consecutive samples one takes from a certain start point. The stride indicates the number of samples the start point is shifted for the next window. A more detailed description of the dataset can be found in the methods and section \ref{methods:data}. 

\subsection{Comparison of different window and stride sizes}

\begin{figure}
     \centering
     \begin{subfigure}[b]{0.49\linewidth}
         \centering
         \includegraphics[width=\textwidth]{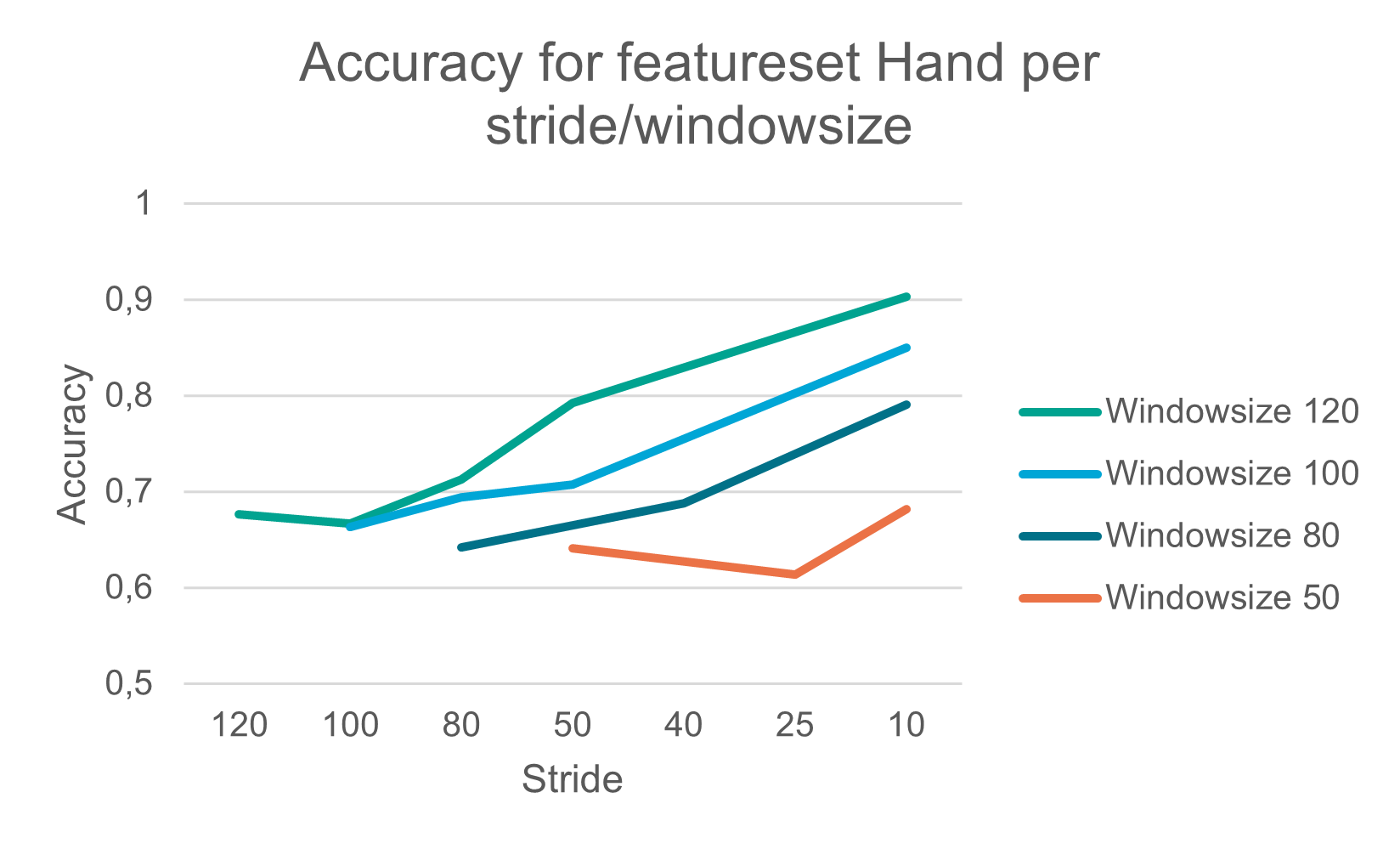}
         \caption{Random split}
         \label{fig:a}
     \end{subfigure}
     \hfill
     \begin{subfigure}[b]{0.49\linewidth}
         \centering
         \includegraphics[width=\textwidth]{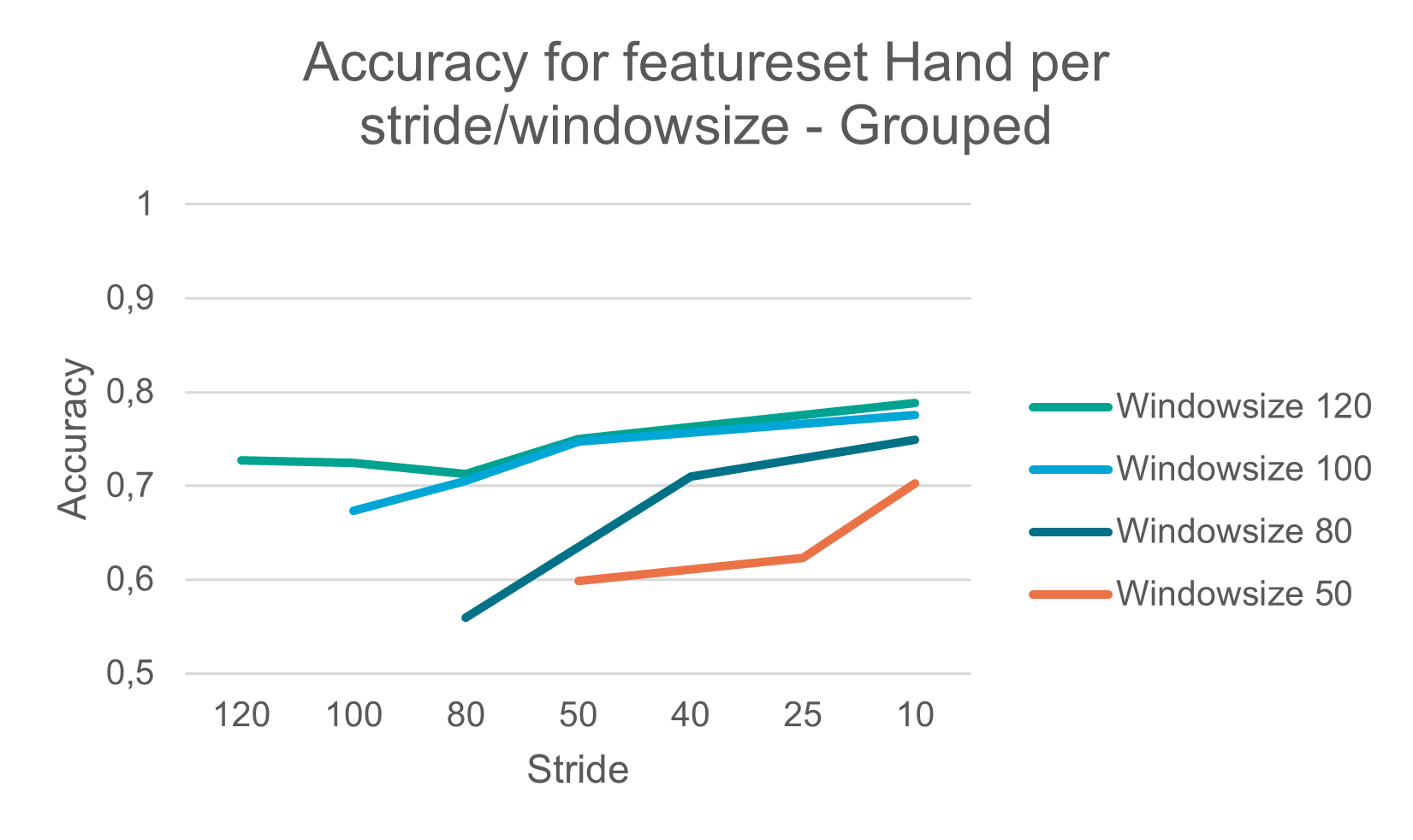}
         \caption{Leave-subject-out}
         \label{fig:b}
     \end{subfigure}
        \caption{Accuracy when training an SVM using (a): a random split (windows of the same subject, both in the train and test set) and (b): when windows of subjects are either in the training or in the test set (leave-subject-out validation procedure). This is inspected for different window and stride sizes. For random splitting, the accuracy increases as the window size increase and stride size decrease, whereas  for the leave-subject-out procedure, the accuracy seems to converge to one point. The achieved accuracies are plotted on the y-axis and the stride sizes are on the x-axis. The different window sizes are represented by different coloured lines. 120 (green), 100 (light blue), 80 (dark blue) and 50 (orange). }
        \label{fig:windowstrideOverlapTrainTest}
\end{figure}

First, we investigate the influence of varying window and stride sizes. For that, we trained a Support Vector Machine (SVM) on windows sizes of 50, 80, 100 and 120 and stride sizes of 10, 25, 40, 50, 80, 100 and 120. A short explanation of the SVM can be found in the methods and appendix \ref{methods:models}. The input data used was the raw time series. We performed the experiment twice. During the first time, we used a train and test set where some windows of a person were in the train set and some were in the test set. The second time we only used a train and test set where all the windows of a person were either in the train or the test set (resulting in a "leave-subject-out" validation procedure). The results for the first two experiments can be found in Figure \ref{fig:windowstrideOverlapTrainTest}.

% \begin{figure}[h]
%     \centering
%     \includegraphics[width=0.5\textwidth]{Images/windowstrideOverlapTrainTest.png}
%     \caption{Achieved accuracy when training an SVM with overlapping train/test set with different window and/or stride size. The accuracy increases as the window size increase and stride size decrease. The achieved accuracies are plotted on the y-axis and the stride sizes are on the x-axis. The different window sizes are represented by different coloured lines. }
%     \label{fig:windowstrideOverlapTrainTest}
% \end{figure}

% \begin{figure}[h]
%   \includegraphics[width=.49\columnwidth]{Images/windowstrideOverlapTrainTest.png}\hfill
%   \includegraphics[width=.49\columnwidth]{Images/windowstrideDistinctTrainTest.png}

%   \caption{Left: Accuracy when training an SVM, when having windows of the same subject, both in the train and test set, for different window and stride sizes. The accuracy increases as the window size increase and stride size decrease. Right: Accuracy when training the SVM when windows of subjects are either in the training or in the test set (leave-subject-out validation procedure) for different window and stride sizes. The accuracy seems to converge to one point. The achieved accuracies are plotted on the y-axis and the stride sizes are on the x-axis. The different window sizes are represented by different coloured lines. 120 (green), 100 (light blue), 80 (dark blue) and 50 (orange). }
%   \label{fig:windowstrideOverlapTrainTest}
% \end{figure}

Both figures clearly illustrate that with every color-coded line, representing different window sizes, the accuracy increases as the stride size decreases. As the stride size decreases, there is an increase in the sample size. However, even though these windows become more dependent (due to larger overlap) with the reduction in stride size, the effectively larger sample size still enhances performance. Moreover, as the window sizes get larger, the accuracy also gets higher. However, there is a difference between the two figures. For the "leave-subject-out" validation procedure, the accuracies seem to converge to one point or at least stabilise, whereas in the random split scenario, the lines show an overall increasing trend.

% \begin{figure}[h]
%     \centering
%     \includegraphics[width=0.5\textwidth]{Images/windowstrideDistinctTrainTest.png}
%     \caption{Achieved accuracy when training an SVM with distinct train/test set with different window and/or stride size. The accuracy seems to converge to one point. The achieved accuracies are plotted on the y-axis and the stride sizes are on the x-axis. The different window sizes are represented by different coloured lines.}
%     \label{fig:windowstrideDistinctTrainTest}
% \end{figure}

\subsection{The effect of clustering subjects}

\begin{figure}[h]
    \centering
    \includegraphics[width=\textwidth]{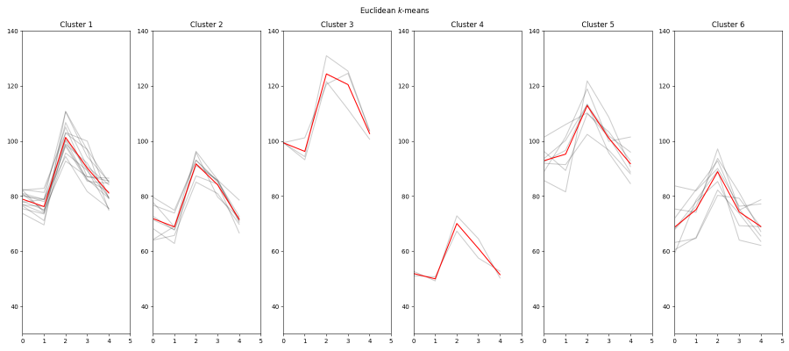}
    \caption{A cluster assignment with the number of clusters equal to 6 using a time series of a subject's mean BPM per activity using the TimeSeriesKmeans clustering procedure\cite{tslearn}. Subplots from left to right represent the six different clusters and the subjects included. Grey lines represent the individual time series and thus represent a single subject. Red lines are the averages of the time series in the cluster. The x-axis shows the different activities numbered from 0 to 4 and the y-axis shows the heart rate in BPM.}
    \label{fig:startClusteringmeanbpm}
\end{figure}

Although the classification performance shown in the previous section is reasonable, it is known that these data show large inter-person variability due to physical differences between subjects \cite{RN98}. To see if more personalised models improve performance, we cluster the subjects based on various metrics. The first metric: the average heart rate (in BPM) of every activity resulted in five values per person. These five values represented a time series of five points in the order as the activities performed: rest, breath, activity, rest, and type. When we clustered these time series per person with different resulting numbers of clusters, a cluster assignment as in Figure \ref{fig:startClusteringmeanbpm} was achieved.

To determine whether there were differences between the cluster groups, we trained an SVM on one cluster while another cluster was used as a testing set. The combinations and the corresponding scores achieved are represented in Table \ref{tab:tablestartClusteringmeanbpm}.

% \begin{table}[h]
% \begin{tabularx}{\columnwidth}{YY}
% \hline
% Train x / Test y & Averaged balanced accuracy \\ \hline
% Train 1 / Test 2 & .7261                     \\
% Train 1 / Test 5 & .4035                     \\
% Train 1 / Test 6 & .5724                     \\
% Train 5 / Test 6 & .3572                     \\
% Train 5 / Test 3 & .4464                    
% \end{tabularx}
\begin{table}[h]
\begin{tabularx}{\columnwidth}{YY}
\hline
Train x / Test y & Averaged balanced accuracy \\ \hline
Train 1 / Test 2 & .73                     \\
Train 1 / Test 5 & .40                     \\
Train 1 / Test 6 & .57                     \\
Train 5 / Test 6 & .36                     \\
Train 5 / Test 3 & .44                    
\end{tabularx}
\caption{Accuracies of training an SVM and using subjects of one cluster as training set and subjects of another cluster as testing set. The numbers indicate the clusters in Figure \ref{fig:startClusteringmeanbpm} counted from left to right. Similar clusters achieve higher accuracy than more dissimilar ones.}
\label{tab:tablestartClusteringmeanbpm}
\end{table}

We can observe in this table that the clusters that look similar (eg. cluster 1 and 2) achieve a cross-cluster better performance than clusters that look more dissimilar (eg. 1 and 5). This suggests that there exists inter-subject variability in this dataset. 

\begin{figure}
     \centering
     \begin{subfigure}[b]{0.49\linewidth}
         \centering
    \includegraphics[width=\textwidth]{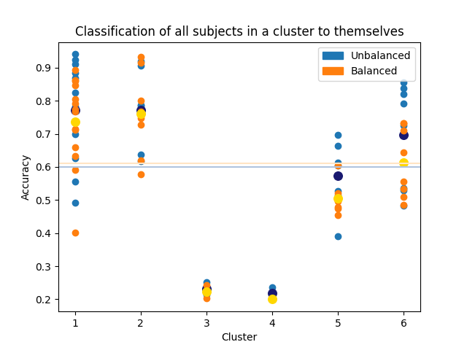}
    \caption{Feature standardization}
    \label{fig:withinclustersFeaturestand}
     \end{subfigure}
     \hfill
     \begin{subfigure}[b]{0.49\linewidth}
         \centering
    \includegraphics[width=\textwidth]{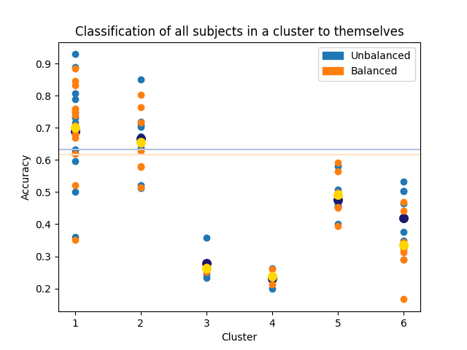}
    \caption{Data standardization}
    \label{fig:withinclustersDatastand}
     \end{subfigure}
        \caption{Results of accuracies within a cluster for the Feature and Data standardization methods when training an SVM with the leave-one-subject out validation procedure. Yellow and dark blue points represent the mean per cluster and horizontal lines represent the performance of the SVM when no clustering is performed. For yellow/orange points, balanced accuracy was used and for light/dark blue, unbalanced/normal accuracy. With Feature standardization (a), three of the four larger clusters have a mean accuracy higher than an SVM trained on all the data. With Data standardization (b), only two of the four larger clusters, have a mean accuracy higher than an SVM trained on all the data. }
        \label{fig:cluster-performance}
\end{figure}

To investigate the existence of variability within a cluster, we trained an SVM on all the data in a cluster except for one subject, which was used for testing. We performed this for every cluster and for every combination inside a cluster. We considered two different standardization methods namely `Feature' and `Data' standardization. In Feature standardization, z-score standardization is applied on the features after windowing and feature generation. In Data standardization, z-score standardization is applied on the original heart rate time per person, whereafter windowing and feature generation is performed. The mean and standard deviation used for standardising the training data are also used for the standardization of the features in the testing data. The results of both methods can be found in Figures \ref{fig:withinclustersFeaturestand} and \ref{fig:withinclustersDatastand}.

% \begin{figure}[h]
%     \centering
%     \includegraphics[width=0.5\textwidth]{Images/withinClustermeanbpmFeaturestand.png}
%     \caption{Results of accuracies within a cluster for the Feature standardization method when training an SVM with the leave-one-subject out validation procedure. Yellow and dark blue points represent the mean per cluster and horizontal lines represent the performance of the SVM when no clustering is performed.. For ellow/orange points, balanced accuracy was used and for light/dark blue, unbalanced/normal accuracy. In three out of the four larger clusters, the mean accuracy within a cluster is higher than an SVM trained on all the data. }
%     \label{fig:withinclustersFeaturestand}
% \end{figure}

% \begin{figure}[h!]
%     \centering
%     \includegraphics[width=0.5\textwidth]{Images/withinClustermeanbpmDatastand.png}
%     \caption{Results of accuracies within a cluster for the Data standardization method when training an SVM with leave-one-subject out testing. Larger yellow and dark blue points represent the mean per cluster and horizontal lines represent the accuracy of the SVM when trained on a distinct train/test set. Light/dark blue represents unbalanced and yellow/orange represents balanced. In two out of the four larger clusters, the mean accuracy within a cluster is higher than an SVM trained on all the data.}
%     \label{fig:withinclustersDatastand}
% \end{figure}

First, these figures show us that the Feature standardization case is performing better. Next, we see that in three out of four (larger) clusters, the average accuracy within a cluster is higher than the SVM when no clustering of subjects is done. Note that clusters 3 and 4 contain an insufficient number of samples to provide an accurate representation. \par
\begin{figure}
     \centering
     \begin{subfigure}[b]{0.49\linewidth}
         \centering
    \includegraphics[width=\textwidth]{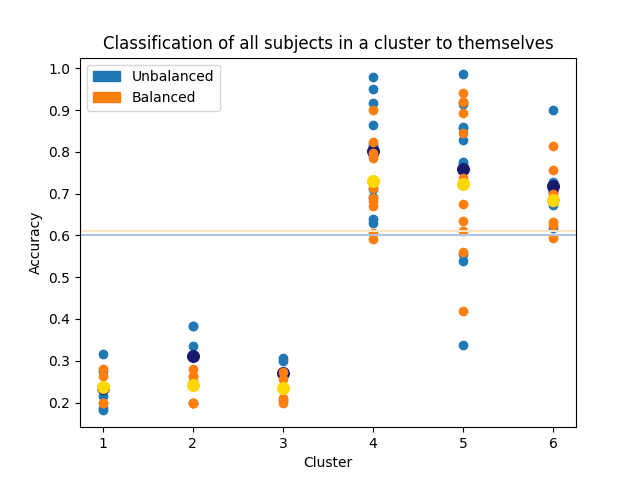}
    \caption{Temporal features}
    \label{fig:withinclustersFeaturestandTemporal}
     \end{subfigure}
     \hfill
     \begin{subfigure}[b]{0.49\linewidth}
         \centering
    \includegraphics[width=\textwidth]{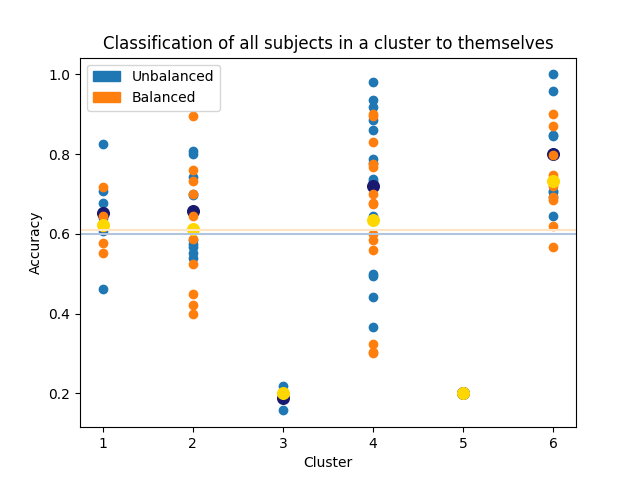}
    \caption{Statistical features}
    \label{fig:withinclustersFeaturestandStat}
     \end{subfigure}
        \caption{Results of accuracies within a cluster for the Feature standardization method using temporal and statistical features and training an SVM with the leave-one-subject out validation procedure. Yellow and dark blue points represent the mean per cluster and horizontal lines represent the performance of the SVM when no clustering is performed. For yellow/orange points, balanced accuracy was used and for light/dark blue, unbalanced/normal accuracy. With temporal features (a), three of the four larger clusters have a mean accuracy higher than an SVM trained on all the data. With statistical features (b), all four clusters have a mean accuracy higher than an SVM trained on all the data. }
        \label{fig:cluster-performances-features}
\end{figure}

Next, we conducted an additional experiment to investigate if the clustering could be improved by using multiple features instead of only the mean heart rate per activity. To test this, we evaluated the within-cluster accuracies using different methods of clustering. The two different methods we investigated were the use of temporal features and the use of statistical features instead of mean heart rate. Statistical and temporal features
are the features generated by TSFEL\cite{barandas2020tsfel}. The results can be seen in Figures \ref{fig:withinclustersFeaturestandTemporal} and \ref{fig:withinclustersFeaturestandStat}.

% \begin{figure}[h!]
%     \centering
%     \includegraphics[width=0.5\textwidth]{Images/accuracyWithinCluster_Apple_SVM_temp.png}
%     \caption{Results of accuracies within a cluster for the Feature standardization method when training an SVM with leave-one-subject out testing and temporal features for clustering. Larger yellow and dark blue points represent the mean per cluster and horizontal lines represent the accuracy of the SVM when trained on a distinct train/test set. In three out of the four larger clusters, the mean accuracy within a cluster is higher than an SVM trained on all the data.}
%     \label{fig:withinclustersFeaturestandTemporal}
% \end{figure}

% \begin{figure}[h!]
%     \centering
%     \includegraphics[width=0.5\textwidth]{Images/accuracyWithinCluster_Apple_SVM_stat.png}
%     \caption{Results of accuracies within a cluster for the Feature standardization method when training an SVM with leave-one-subject out testing and statistical features for clustering. Larger yellow and dark blue points represent the mean per cluster and horizontal lines represent the accuracy of the SVM when trained on a distinct train/test set. In all of the four larger clusters, the mean accuracy within a cluster is higher than an SVM trained on all the data.}
%     \label{fig:withinclustersFeaturestandStat}
% \end{figure}

These figures show that the statistical features are better for clustering than the temporal features. In all large clusters, it achieves better performance than the SVM trained when no clustering is performed. In the temporal case, this is only 3 out of 4 just like with the mean BPM clustering method. 

To demonstrate that it can also help with previously unseen samples, we conducted several additional experiments using the leave-subject-out procedure. We used the training set for generating the clustering model and cluster assignment, as well as to train a model for each cluster. The test set was used in two different ways. The first approach was per-window classification. With this approach, a window of a test subject was obtained, the corresponding cluster was determined, and the model associated with that cluster was used to classify the window. The results of this approach can be seen in the first column of Table \ref{tab:withinclustersMajority}.

\begin{table}[h!]
\begin{tabularx}{\columnwidth}{@{}ZZZ@{}}
\toprule
    & Per-window  & Per-subject \\ \midrule
6 clusters        & .46       & .74              \\ \midrule
5 clusters        & .50       & .63              \\ \midrule
4 clusters        & .56       & .75              \\ \midrule
3 clusters        & .68       & .72              \\ \bottomrule
\end{tabularx}
\caption{Achieved classification accuracies when using different numbers of resulting clusters and clustering techniques to find a cluster model for activity prediction. Grouping the subjects in 4 clusters and using the per-subject method achieves the highest accuracy.}
\label{tab:withinclustersMajority}
\end{table}

The second approach was to apply the same personalised classifier to all windows of one test subject, the per-subject approach. To do so, we used the clustering model to determine to which cluster each single window of a test subject belongs to. After this, the cluster with the highest number of assigned windows was used to obtain the model for classifying all windows of a specific subject. The result of this experiment is presented in the second column of Table \ref{tab:withinclustersMajority}. The per-subject method achieves higher accuracies than the per-window classification method. It achieves an accuracy of 0.71 while the per-subject method with 4 clusters achieves 0.75. In the next paragraph, we delve more into the differences in prediction between both methods, rather than solely examining the achieved accuracies. 

\begin{figure}
     \centering
     \begin{subfigure}[b]{0.49\linewidth}
         \centering
    \includegraphics[width=\textwidth]{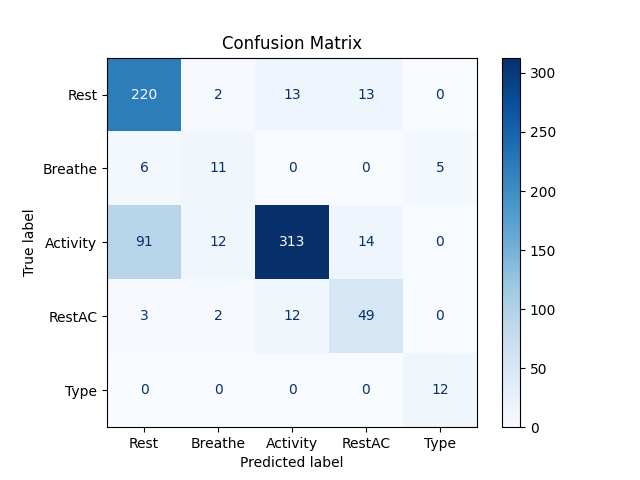}
    \caption{Per-window}
    \label{fig:cm-per-window}
     \end{subfigure}
     \hfill
     \begin{subfigure}[b]{0.49\linewidth}
         \centering
    \includegraphics[width=\textwidth]{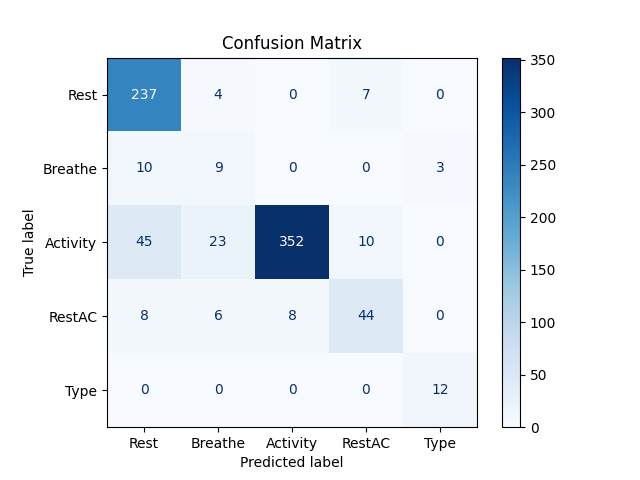}
    \caption{Per-subject}
    \label{fig:cm-per-subject}
     \end{subfigure}
        \caption{Confusion matrices for the per-window  and the per-subject approach. The true/actual labels are shown on the vertical axis and the predicted labels are on the horizontal axis. The biggest difference can be seen in the predictions of the Rest and Activity class.}
        \label{fig:Confusionmatrices}
\end{figure}

The confusion matrices of the two methods can be found in Figure \ref{fig:Confusionmatrices}. We can see a prominent difference within misclassifications occuring between the two largest classes (Rest and Activity). The per-subject approach exhibits much less misclassifications between these two classes compared to the per-window method. Depending on the application, misclassifications between these very different classes is more severe than misclassifications between similar classes (Rest vs Breathe and RestAC). Overall, the per-subject model performs better.
% \begin{figure}[h]
%   \includegraphics[width=.49\columnwidth]{Images/SVM_ConfusionMatrix_Apple_Window_ClusterGroup_train_test_120_10_Temporal_bal.png}\hfill
%   \includegraphics[width=.49\columnwidth]{Images/confusionmatrix_clstr4.png}

%   \caption{Confusion matrices for the per-window  and the per-subject approach. The true/actual labels are shown on the vertical axis and the predicted labels are on the horizontal axis. The biggest difference can be seen in the predictions of the Rest and Activity class. }
%   \label{fig:Confusionmatrices}
% \end{figure}
% The second point that we can notice is the difference in the classification of the Activity class. In both cases, the Activity class is misclassified as Rest, Breathe or RestAC. However, the clustering model performs the classification better by having fewer mispredictions in general and less misprediction in Rest, which is the least probable among the three classes that are occasionally predicted instead of Activity. In general, we find that clustering primarily helps with reducing the misclassification of the Rest and Activity classes.

\subsection{Deep learning with handcrafted features}

Current research mostly focuses on deep-learning networks for feature extraction and classification. Especially for heart rate variability analysis, there exist some standard features for measuring the variability. Although they are typically manually constructed, and therefore often interpretable, one may wonder if these measures capture all information needed for health diagnosis or activity recognition. In this section, the statistical and temporal features used in earlier experiments are incorporated into a Deep Learning method to investigate whether incuding these handcrafted (HC) features can improve activity classification. 

\begin{figure}
     \centering
     \begin{subfigure}[b]{0.49\linewidth}
         \centering
    \includegraphics[width=\textwidth]{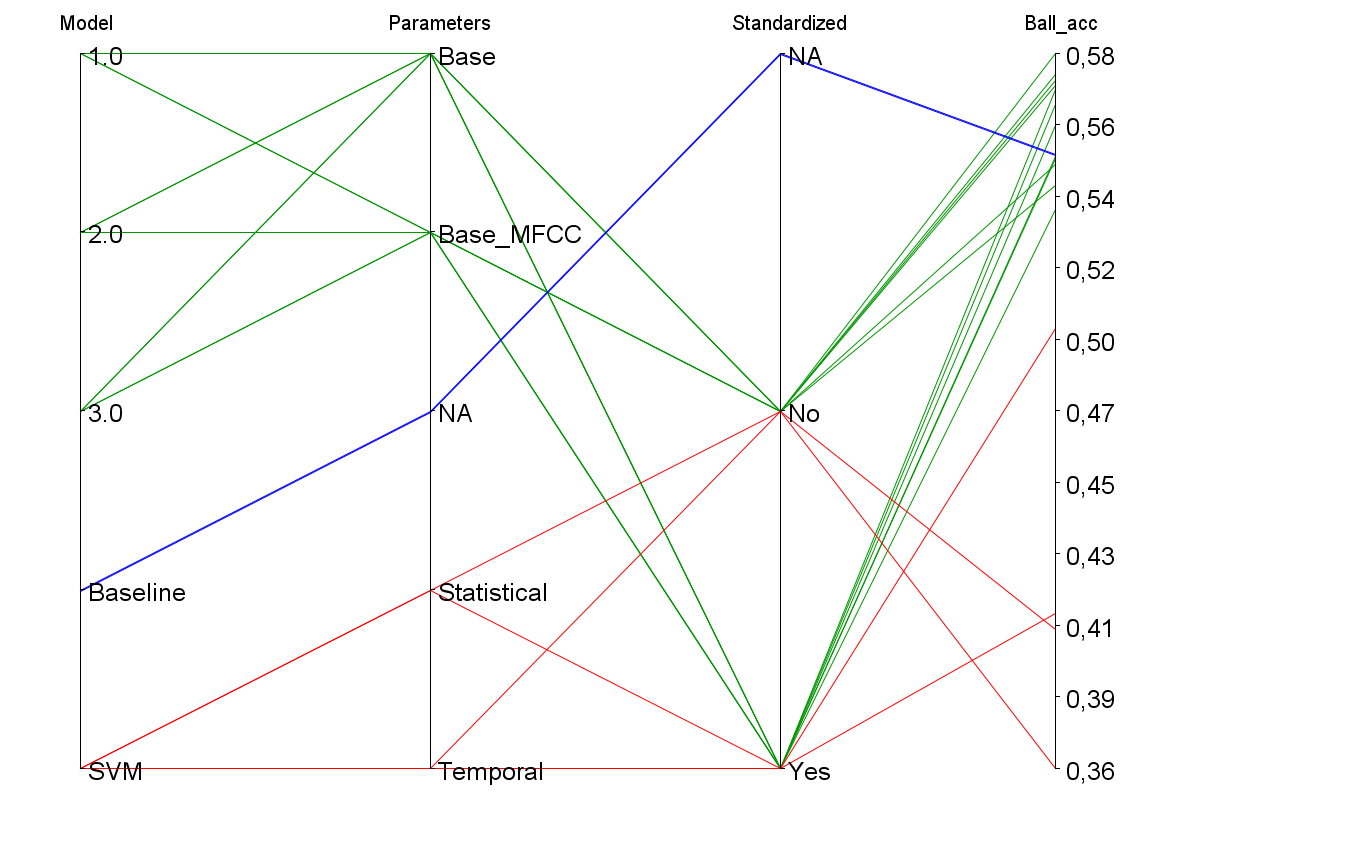}
    \caption{DL model base and MFCC features vs SVM statistical and temporal features.}
    \label{fig:flow1}
     \end{subfigure}
     \hfill
     \begin{subfigure}[b]{0.49\linewidth}
         \centering
    \includegraphics[width=\textwidth]{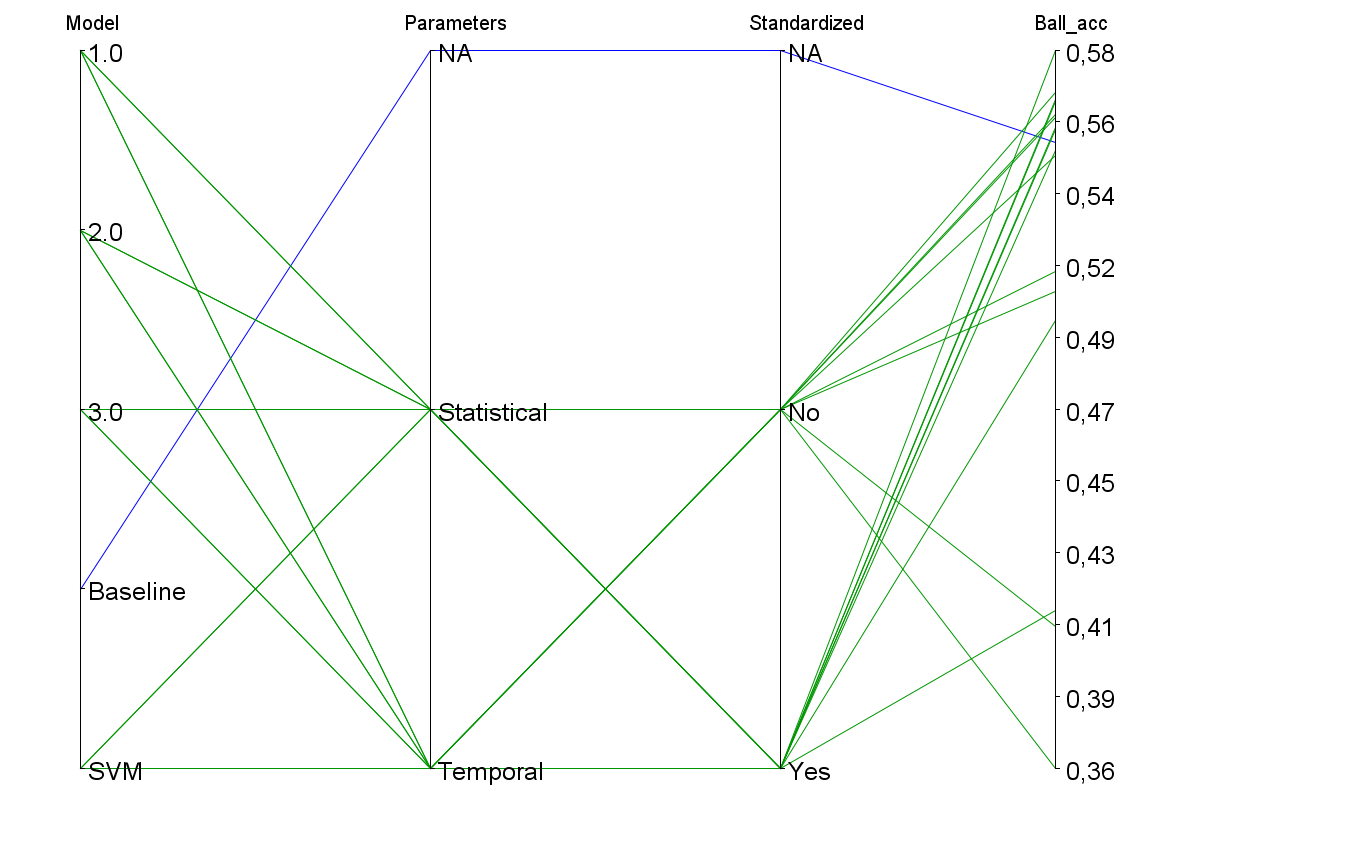}
    \caption{DL vs SVM, both using statistical and temporal features.}
    \label{fig:pcPlotTempStat}
     \end{subfigure}
        \caption{Accuracies of different DL models with base HC features compared to DL baseline and SVM. The blue line is the baseline DL network that gets as input the raw standardized data. The red lines are the runs with an SVM and the green lines are the runs with the proposed DL networks.  It can be seen that all the DL models outperform the SVM and most DL models with HC features outperform the baseline. The best performing configuration was achieved using model 2. Models 1, 2, and 3 represent the three different DL models, which are explained in the method section \ref{methods:models}. Models 1 and 3 make use of late integration and model 2 of early integration of the DL and HC features. The parameters column represents which feature set is used. The base feature set represents the basic HC features like max, min, mean, std and means of different (first and second-order) derivatives on the heart rate values in a time series window. Base and MFCC \cite{RN124} represent the feature set where there are all the base parameters plus MFCC features. Statistical and temporal features are the features generated by TSFEL \cite{RN125}. The column standardized indicates if the HC features are calculated on the standardized input or not. The raw heart rate time series data is always standardized. When we talk about standardized or non-standardized HC features in the next sections, we mean the features calculated on a standardized or non-standardized input.}
        \label{fig:pcPlotBaselineSVM}
\end{figure}

First of all, we compared the performance of the SVM models with the deep learning models (convolutional neural networks) with and without HC features. The results can be found in Figure \ref{fig:pcPlotBaselineSVM} and illustrates that the addition of HC features results in an increase in balanced accuracies in comparison to the DL baseline model in certain instances. Additionally, the top four accuracies are achieved without standardizing the HC features. Furthermore, every DL model outperformed the SVM. Lastly, the second DL model which makes use of HC features and a window size of 80 and stride of 10, performs the best compared to all the models and configurations.

% \begin{figure}[h!]
%     \centering
%     \includegraphics[width=0.8\textwidth]{Images/ParralelCoordinatesBaseSVMv2.png}
%     \caption{Accuracies of different DL models with base HC features compared to DL baseline and SVM. The blue line is the baseline DL network that gets as input the raw standardized data. The red lines are the runs with an SVM and the green lines are the runs with the proposed DL networks.  It can be seen that all the DL models outperform the SVM and most DL models with HC features outperform the baseline. Models 1, 2, and 3 represent the three different DL models, which are explained in the method section \ref{methods:models}. Models 1 and 3 make use of late integration and model 2 of early integration of the DL and HC features. The parameters column represents which feature set is used. The base feature set represents the basic HC features like max, min, mean, std and means of different (first and second-order) derivatives on the heart rate values in a time series window. Base and MFCC \cite{RN124} represent the feature set where there are all the base parameters plus MFCC features. Statistical and temporal features are the features generated by TSFEL \cite{RN125}. The column standardized indicates if the HC features are calculated on the standardized input or not. The raw heart rate time series data is always standardized. When we talk about standardized or non-standardized HC features in the next sections, we mean the features calculated on a standardized or non-standardized input.}
%     \label{fig:pcPlotBaselineSVM}
% \end{figure}

Next, we investigated the usage of temporal and statistical features as an alternative to the base set of HC features. Each DL model was trained with temporal or statistical features and with or without standardization. The outcomes can be found in Figure \ref{fig:pcPlotTempStat}. Among the top eight accuracies, five configurations employed standardized input. While in the previous experiment with the base features, the non-standardized HC features performed better. In addition to this, we combined both the statistical and temporal features into a single feature set, resulting in a slight improvement in performance to 58.84 \% accuracy. Similarly, in this experiment, the standardized HC feature set worked better than the non-standardized one.

% \begin{figure}[h!]
%     \centering
%     \includegraphics[width=0.8\textwidth]{Images/ParralelCoordinatesTempStatSVM.png}
%     \caption{Accuracies of different DL models with temporal and statistical features compared to DL baseline and SVM. The DL baseline is shown in blue, the SVM in red and the DL models with HC features in green. Again, it can be noticed that all the DL models outperform the SVM and some DL models with HC features outperform the baseline. }
%     \label{fig:pcPlotTempStat}
% \end{figure}

\begin{figure}[h!]
    \centering
    \includegraphics[width=0.5\textwidth]{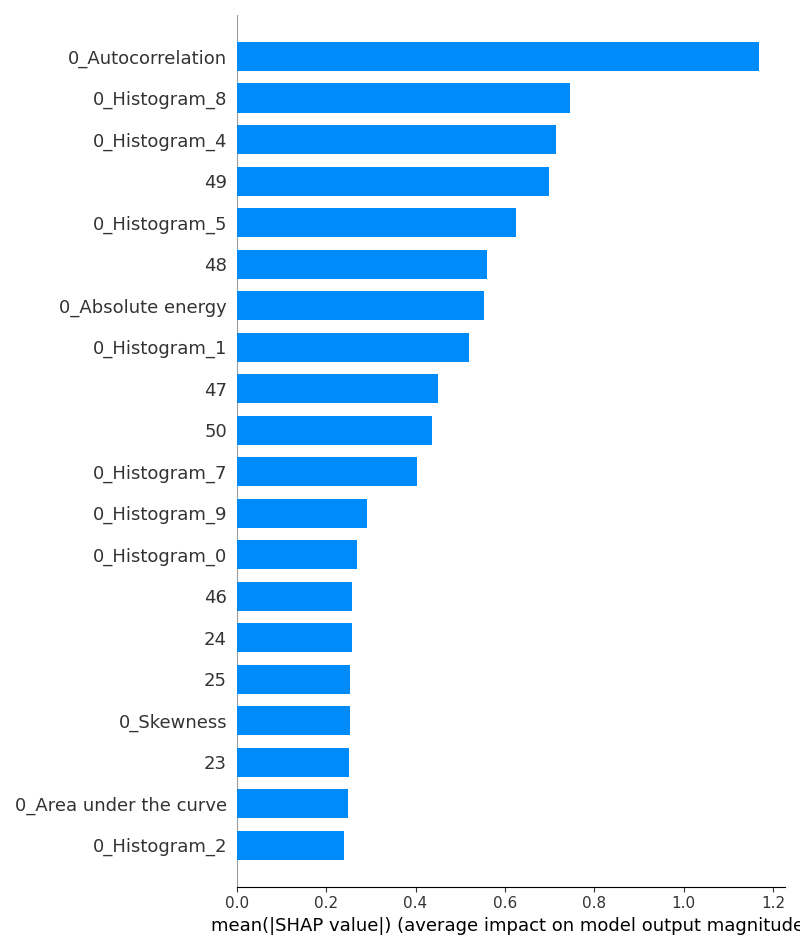}
    \caption{Resulting top 20 SHAP values of 3rd DL model with the addition of temporal + statistical features. On the x-axis the SHAP value is shown and on the y-axis the feature. Features starting with `0\_' indicate HC features, and numbers indicate the index of a heart rate value in the time series window input.}
    \label{fig:shap3rdmodel}
\end{figure}

Besides solely examining the accuracies, it is relevant to investigate whether the HC features were indeed utilized by the DL model. To this extent, we used SHAP values to see how important the HC features are in addition to the raw input data \cite{lundberg2017unified}. Figure \ref{fig:shap3rdmodel} depicts the top 20 SHAP values with the highest importance. As we can see, the highest SHAP values correspond to an HC feature: 0\_Autocorrelation. Another observation is that primarily the heart rate values in the middle or the end of a window input are important (using a window size of 50). 

\subsection{Misclassification with DL models}
\begin{figure}[h!]
    \centering
    \includegraphics[width=0.5\textwidth]{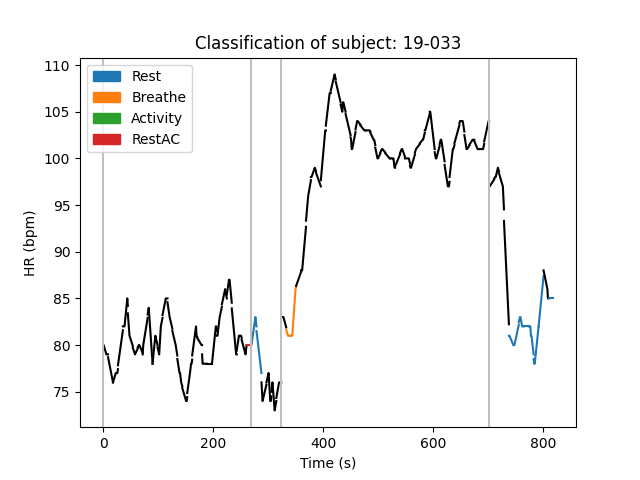}
    \caption{Plot which visualizes the misclassification of the DL model. Black lines correspond to rightly predicted classes and a coloured section indicates a misprediction of a specific class. The colour corresponds to the class that is mispredicted. The grey vertical bars correspond to the change in activity according to the labeling of the data. }
    \label{fig:plotDLMisclassification}
\end{figure}

To interpret the predictions of the DL model with HC statistical and temporal features, a time series is plotted where the line colour indicates if the prediction is correct or incorrect, and to what class it is misclassified. An example is shown in Figure \ref{fig:plotDLMisclassification}. Most of the misclassifications happen after a change of class (for instance, at t=270, where the class changes from Breathe to Activity). Intuitively this makes sense as the heart rate measurements do not immediately change during activity change and therefore it is difficult to predict the activity accurately. 

\newpage
\section{Discussion \& Conclusion}
\label{sec:discusion}
We looked into the classification of activity using only heart rate time series. Results show that there seems to be a relation between the optimal window and stride size for classification: Higher window sizes and smaller strides correspond to higher accuracies. In this case, larger window sizes reveal distinct activity patterns that aren't discernible at smaller sizes. The smaller stride sizes ensure that the model is exposed to many time-shifted variations of an activity pattern, during training. This ensures that the model is better prepared to recognize any such variations that might be present during testing.
Because of the large variety in characteristics of wearables and persons (inter-device and inter-subject variability), it is challenging to build one single prediction model that works on everyone. We have shown that
in the context of activity classification in heart rate time series, it is helpful to group similar subjects together and thus creating semi-personalised models, for each separate group of alike persons. This improvement is particularly evident when using multiple windows from a subject to assign the person to a specific group. We believe that this is important to take into consideration when investigating more challenging tasks like heart disease classification.
Furthermore, we have shown that feeding a deep learning model with handcrafted features improves performance when classifying heart rate time series. By adding features manually, the network learns to find patterns in the time series itself but also uses some of the given HC features to make a decision. This result supports the finding by Eltras et al.\cite{RN106}, where they use a DL network for feature extraction and concatenate the learned features with handcrafted features. Here, the best performance was achieved by concatenating the raw features with handcrafted features before inputting to the DL network.

Due to the limited number of subjects available, we did not have enough subjects in some situations to do a proper train/test split within a cluster, which was a common limitation for all the points mentioned above.  This is particularly true for the the semi-personalised models. For those  that were based on enough subjects, a positive increase in performance was shown. Another limitation of our work is that the experiments have only been conducted with one dataset. Finally, other normalization techniques can be explored to even further reduce the variability among the subjects, for example incorporating (meta)data like age or fitness.

% \section{Conclusion}
% \label{sec:conclusionfuturework}

% The aim of this paper was to find out if the heart rate time series could be used for the classification of different activities using the BigIdeasLab\_STEP dataset. 
% First of all windowed data were used to predict which activity a person was performing. The results showed that there were significant differences between the heart rate time series of different subjects due to morphological differences. This resulted in a better performance when creating classifiers for groups of subjects with similar characteristics. Furthermore, when using deep learning networks an improvement was seen when adding handcrafted features in addition to the raw input, instead of only using the raw heart rate time series. Moreover, standardization based on the resting heart rate per person was more effective than standardization based on the entire population. All in all, heart rate time series can be used for the classification task of predicting a specific activity, but the subject variability should be taken into account by utilizing techniques like grouping or normalizing. 

\appendix
\section{Methods}
\label{sec:methology}

\subsection{BigIdeasLab\_STEP}
\label{methods:data}

In this paper we used the BigIdeasLab\_STEP dataset from PhysioNet \cite{BigIdeasLabDS}. This dataset includes data from 53 participants and was recorded in July-August 2019. The age of the participants ranged from 18 to 54. Each person needed to perform three study protocol rounds with different types of wearables. One study protocol round consisted of five activities in the following order: 
\begin{enumerate}
    \item Seated rest (4 min)
    \item Paced deep breathing (1 min)
    \item Physical activity (5 min)
    \item Seated rest (~2 min)
    \item Typing (1 min)
\end{enumerate}

In the experiment, every person wore all the available devices spread over multiple rounds, capturing different amounts of samples. Round 1:Empatica E4 ($N\approx 140K$), Apple Watch 4($N\approx13K$). Round 2:Fitbit Charge 2($N\approx11K$). Round 3: Garmin Vivosmart 3($N\approx37K$), Xiaomi Miband($N\approx21K$) and Biovotion Everion($N\approx161K$). During the whole experiment, the participant always wore a Bittium Faros 180 ECG device($N\approx221K$) as a reference. 

The dataset consists of a synchronised heart rate value in bpm between the smartwatch and the ECG device. Moreover, it is annotated with one of the five activities the person is performing. In the dataset, this is denoted by the labels Rest, Breathe, Activity, Rest after Activity (RestAC) and Type. In the experiments, only the heart rate data is used of the Apple Watch because of its strong correlation with the heart rate time series of the ECG ground truth in comparison with the other wearables.

\subsection{Classification models}
\label{methods:models}

In several experiments, we used a support vector machine (SVM). An SVM tries to maximize the margin between two classes. The SVM maximizes the generalization of a model \cite{cervantes2020comprehensive}. For multiclass classification one can use multiple binary SVMs. Two of the methods used for this are One-against-all and one-against-one \cite{mayoraz2006support}. For the experiments, we used the implementation provided by scikit-learn, which uses the one-against-one method \cite{scikit-learn}.

In addition to the other experiments, we researched the influence of the addition of handcrafted (HC) features with deep learning models. To investigate this, three different DL models with the addition of HC features were used alongside a DL baseline. All of the DL models started with a 1-D convolution and had three or four fully connected layers. 

\begin{figure}[h!]
  \centering
  \begin{tabular}{@{}c@{}}
    \includegraphics[width=.5\textwidth]{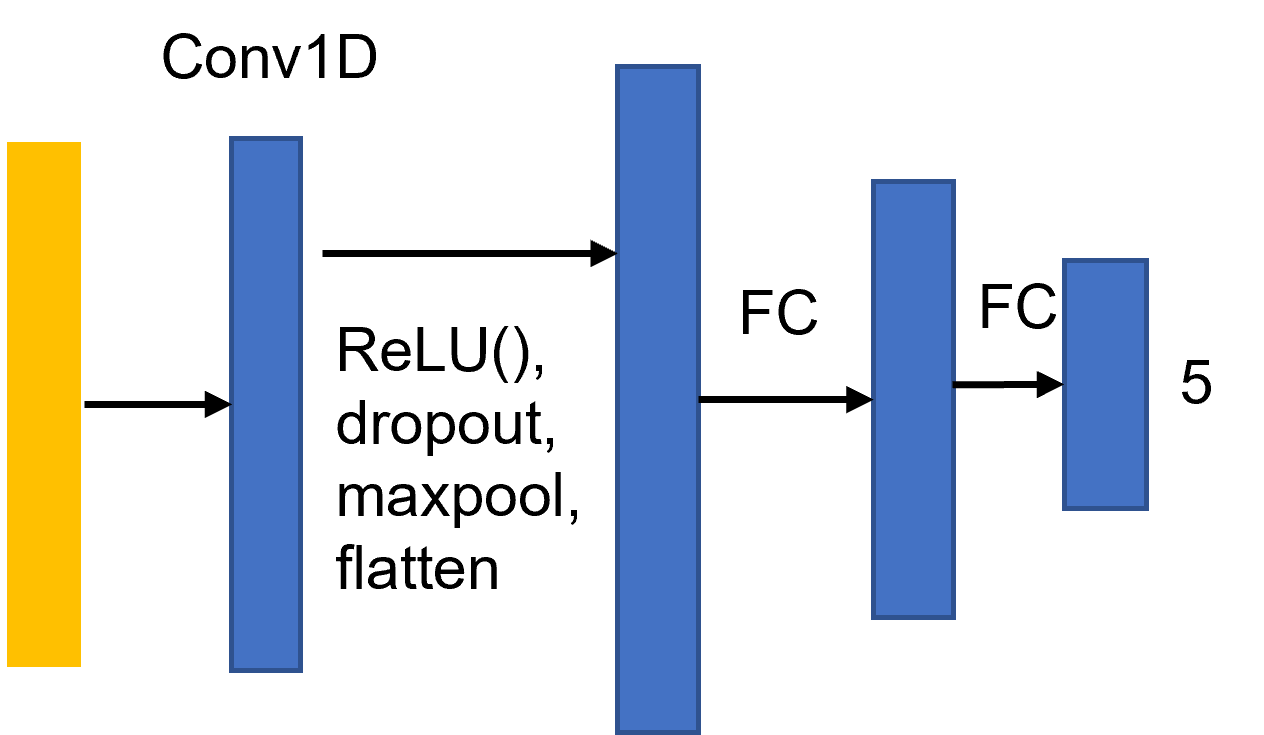} \\[\abovecaptionskip]
    \small (a) baseline network
  \end{tabular}

  \vspace{\floatsep}

  \begin{tabular}{@{}c@{}}
    \includegraphics[width=.5\textwidth]{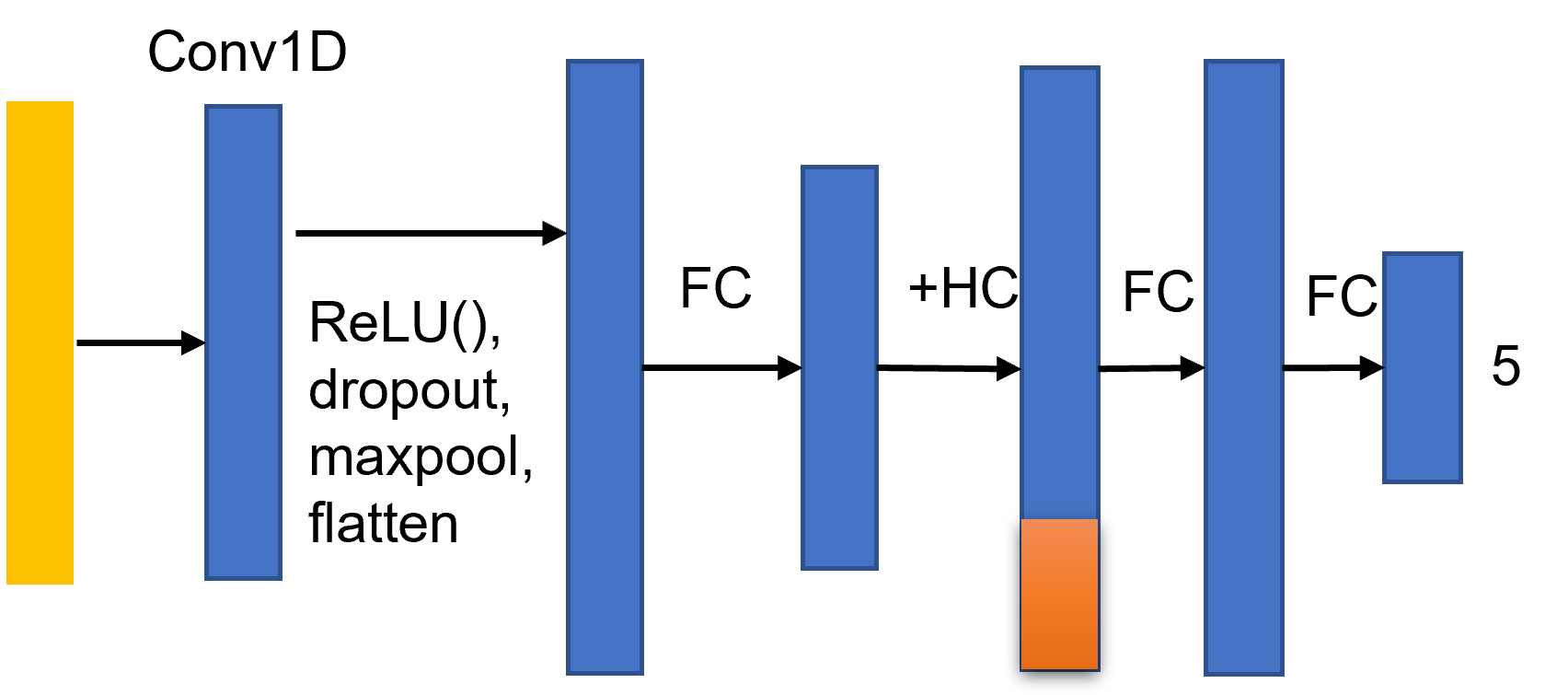} \\[\abovecaptionskip]
    \small (b) First network
  \end{tabular}

  \caption{High-level overview of the internal working of the baseline DL model(a) and the first DL model (b) that make use of HC features. Yellow represents the raw input sequence and orange represents the HC features.}\label{fig:DLnetworksbase1}
\end{figure}

The baseline model starts with a 1-D convolution where the raw sequence input will be processed. Next, it goes through a ReLU, Dropout and Max pooling layer and finally, a flatten layer. After the flattening, it is processed by a fully connected layer, followed by a ReLU and a last fully connected layer to bring the output dimension to the required number of classes. A high-level graphical overview can be seen in Figure \ref{fig:DLnetworksbase1}a.

The first model is highly similar to the baseline model but it adds an extra layer between the last two fully connected layers. So after the first fully connected layer after flattening, the model adds the HC features to the output of this layer. Next, it processes through another fully connected layer and thereafter it goes through the last fully connected layer. This layer ensures that it ends with the correct dimension. A simple graphical representation can be found in Figure \ref{fig:DLnetworksbase1}b. 

\begin{figure}[h!]
  \centering
  \begin{tabular}{@{}c@{}}
    \includegraphics[width=.5\textwidth]{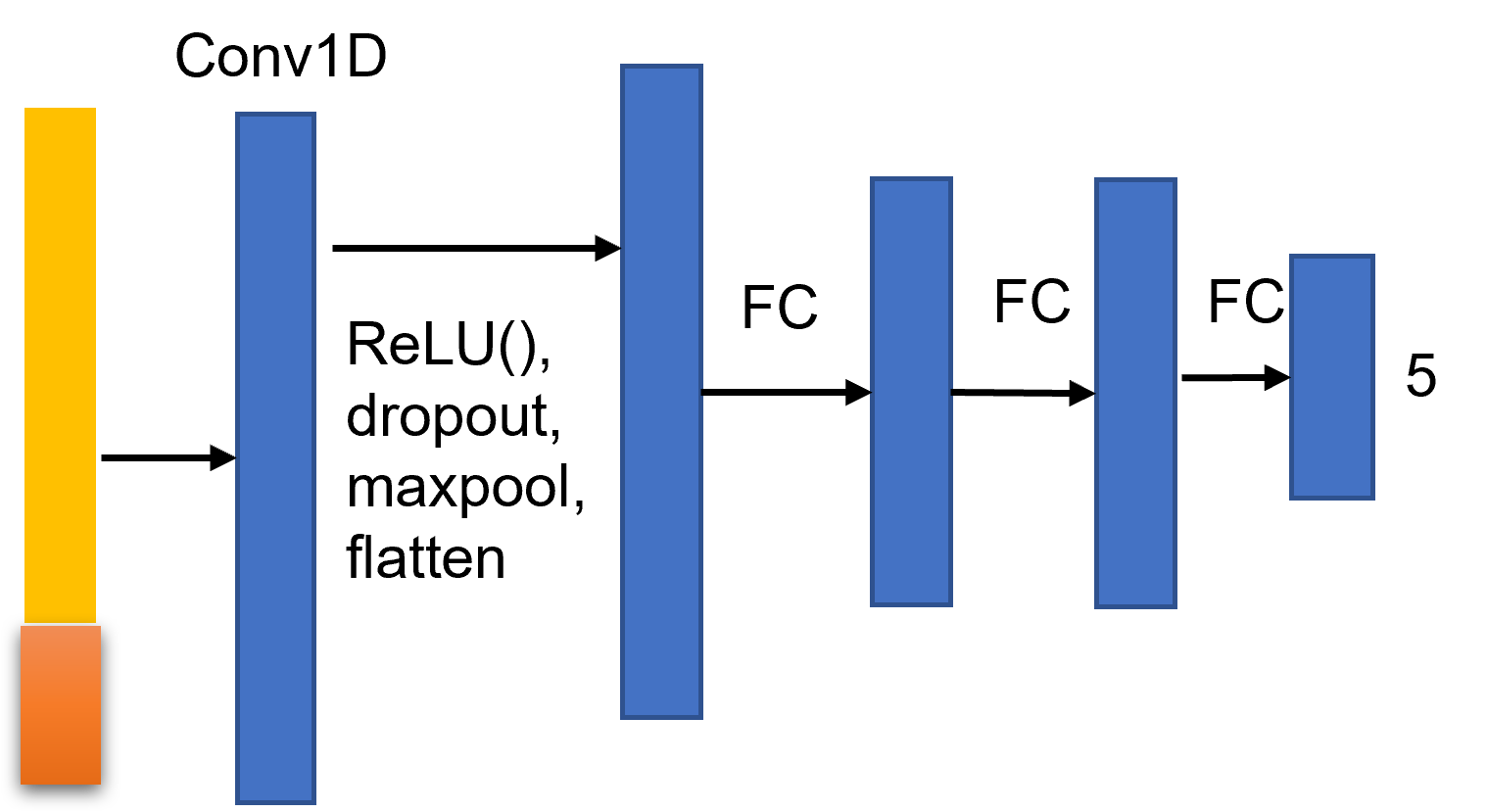} \\[\abovecaptionskip]
    \small (a) Second network
  \end{tabular}

  \vspace{\floatsep}

  \begin{tabular}{@{}c@{}}
    \includegraphics[width=.5\textwidth]{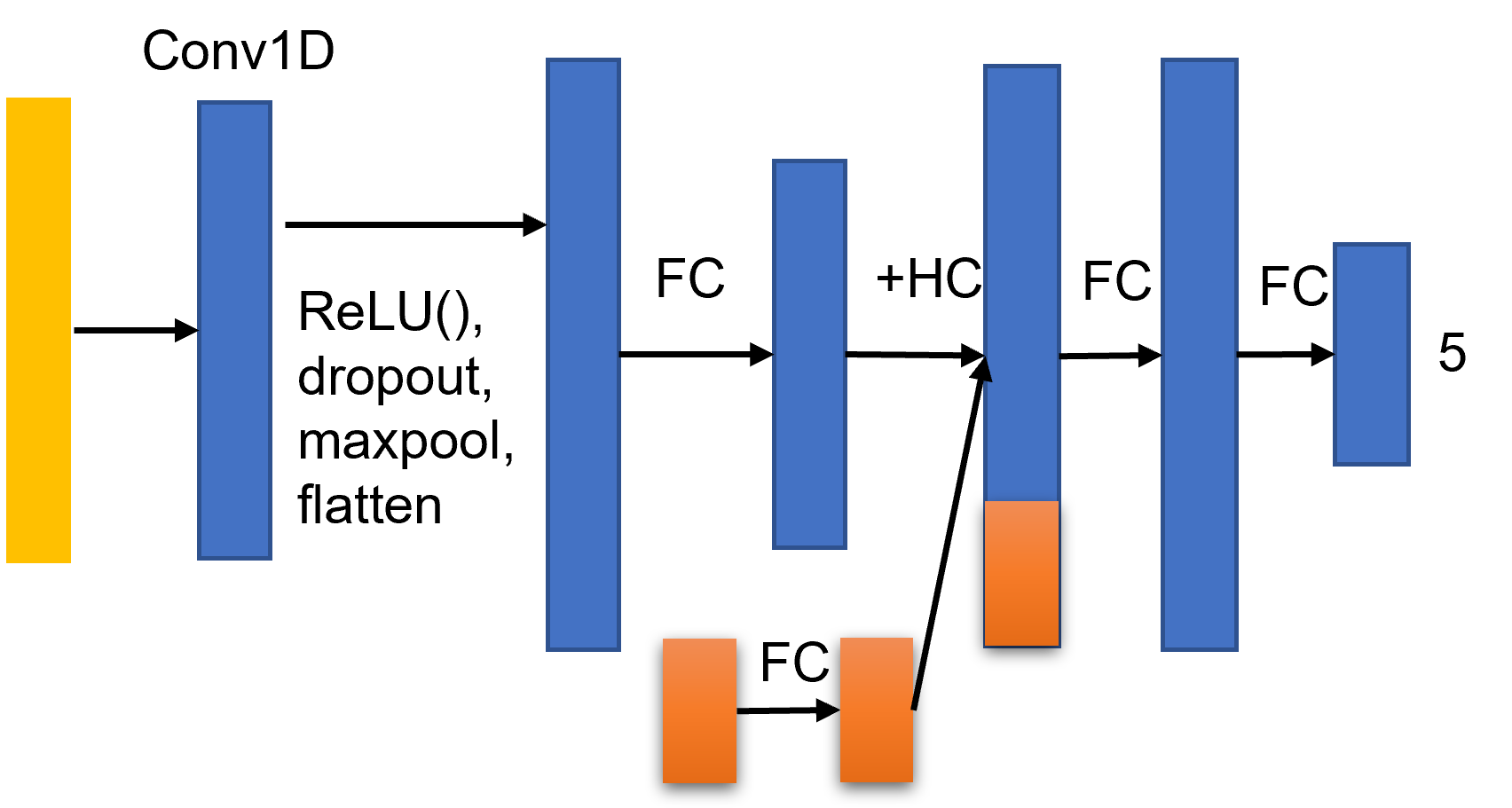} \\[\abovecaptionskip]
    \small (b) Third network
  \end{tabular}

  \caption{High-level overview of the internal working of the second(a) and third(b) DL model that makes use of HC features. Yellow represents the raw input sequence and orange represents the HC features.}\label{fig:DLnetworks23}
\end{figure}

The second model is integrating the HC features directly at the beginning of the DL model. This is achieved by concatenating the HC features with the raw time series input. This results in a larger input vector than with the previous model. The third model is very similar to the first one but with one addition. Instead of adding the HC features directly to the output of the fully connected layer, the HC features first go through a fully connected layer and this output is connected to the output of the first fully connected layer of the model. A graphical representation of both models can be found in Figure \ref{fig:DLnetworks23}.

\clearpage
\bibliographystyle{splncs04}
\bibliography{references}

\begin{thebibliography}{10}
\bibitem{RN131}
Maguire, D., Frisby, R.: Comparison of feature classification algorithm for activity recognition based on accelerometer and heart rate data. In: 9th. IT \& T Conference. p.~11
\end{thebibliography}




% LNCS Homepage, \url{http://www.springer.com/lncs}. Last accessed 4
% \end{thebibliography}
\end{document}